\documentclass{article}

\pdfoutput=1

\PassOptionsToPackage{numbers, compress}{natbib}

\usepackage[american]{babel}
\usepackage{booktabs}
\usepackage{array}
\usepackage{caption}
\usepackage{adjustbox}
\usepackage{mathtools}
\usepackage[preprint]{neurips_2024}

\usepackage[utf8]{inputenc} %
\usepackage[T1]{fontenc}    %
\usepackage{booktabs}       %
\usepackage{amsfonts}       %
\usepackage{nicefrac}       %
\usepackage{microtype}      %

\usepackage{enumitem}
\usepackage{multirow}
\usepackage{adjustbox}
\usepackage{pifont}%
\usepackage{colortbl}
\usepackage{nicefrac}       %
\usepackage{microtype}      %

\usepackage{algorithm}
\usepackage{algcompatible}
\usepackage{subcaption}
\usepackage{amsmath}
\usepackage{listings}
\usepackage{wrapfig}
\usepackage{xcolor}

\newlength\savewidth

\definecolor{codeblue}{rgb}{0.25, 0.5, 0.5}
\definecolor{codekw}{rgb}{0.35, 0.35, 0.75}
\lstdefinestyle{Pytorch}{
    language         = Python,
    backgroundcolor  = \color{white},
    basicstyle = \fontsize{8.0pt}{9pt}\selectfont\ttfamily\bfseries,
    columns          = fullflexible,
    breaklines       = true,
    captionpos       = b,
    commentstyle     = \fontsize{4pt}{4pt}\color{codeblue},
    keywordstyle     = \fontsize{4pt}{4pt}\color{codekw},
    morekeywords     = {with,scatter_,norm,sort},
}

\title{NoRA: Nested Low-Rank Adaptation for Efficient Fine-Tuning Large Models}

\author{
$\text{Cheng Lin}^{1,2 \dag}$, 
$\text{Lujun Li}^{1 \dag}$, 
$\text{Dezhi Li}^{1,3}$, 
$\text{Jie Zou}^{2}$, 
$\textbf{Wei Xue}^{1 *}$, 
$\textbf{Yike Guo}^{1 *}$ \\
$^1$HKUST \quad 
$^2$UESTC \quad 
$^3$SEU 
\thanks{Corresponding authors\{weixue,yikeguo\}@ust.hk \\
$\dag$ Equal contribution \{linchengtech,lilujunai\}@gmail.com. Work in progress, revisions ongoing.}
}
\vspace{-2.5em}
\begin{document}

\maketitle

\begin{abstract}
In this paper, we introduce Nested Low-Rank Adaptation (NoRA), a novel approach to parameter-efficient fine-tuning that extends the capabilities of Low-Rank Adaptation (LoRA) techniques. Vanilla LoRA overlooks pre-trained weight inheritance and still requires fine-tuning numerous parameters. To addresses these  issues,  our NoRA adopts a dual-layer nested structure with Singular Value Decomposition (SVD), effectively leveraging original matrix knowledge while reducing tunable parameters. Specifically, NoRA freezes the outer LoRA weights and utilizes an inner LoRA design, providing enhanced control over model optimization. This approach allows the model to more precisely adapt to specific tasks while maintaining a compact parameter space.  By freezing outer LoRA weights and using an inner LoRA design, NoRA enables precise task adaptation with a compact parameter space. Evaluations on tasks including commonsense reasoning with large language models, fine-tuning vision-language models, and subject-driven generation demonstrate NoRA's superiority over LoRA and its variants.  Code will be released upon acceptance.

\end{abstract}

\section{Introduction}
In recent years, LLMs~\cite{zhao2023survey} have set new performance benchmarks in natural language processing~\cite{touvron2023llama} (NLP) and related fields~\cite{zhou2020evaluating,sap2020commonsense,lin2024sdxl}, yet their substantial size introduces significant challenges in training and adaptation. The high computational and storage demands of comprehensive fine-tuning make it impractical in resource-limited environments. To address this, Parameter-Efficient Fine-Tuning~\cite{ding2023parameter, han2024parameter} (PEFT) techniques have been developed, focusing on fine-tuning a reduced subset of model parameters to optimize model adaptability. Among these techniques, LoRA~\cite{hu2021lora} stands out as a prominent method that mitigates the extensive resource requirements of full fine-tuning~\cite{lv2023full} by introducing trainable low-rank matrices into the model architecture.

\begin{figure*}
    \centering
    \includegraphics[width=0.95\linewidth]{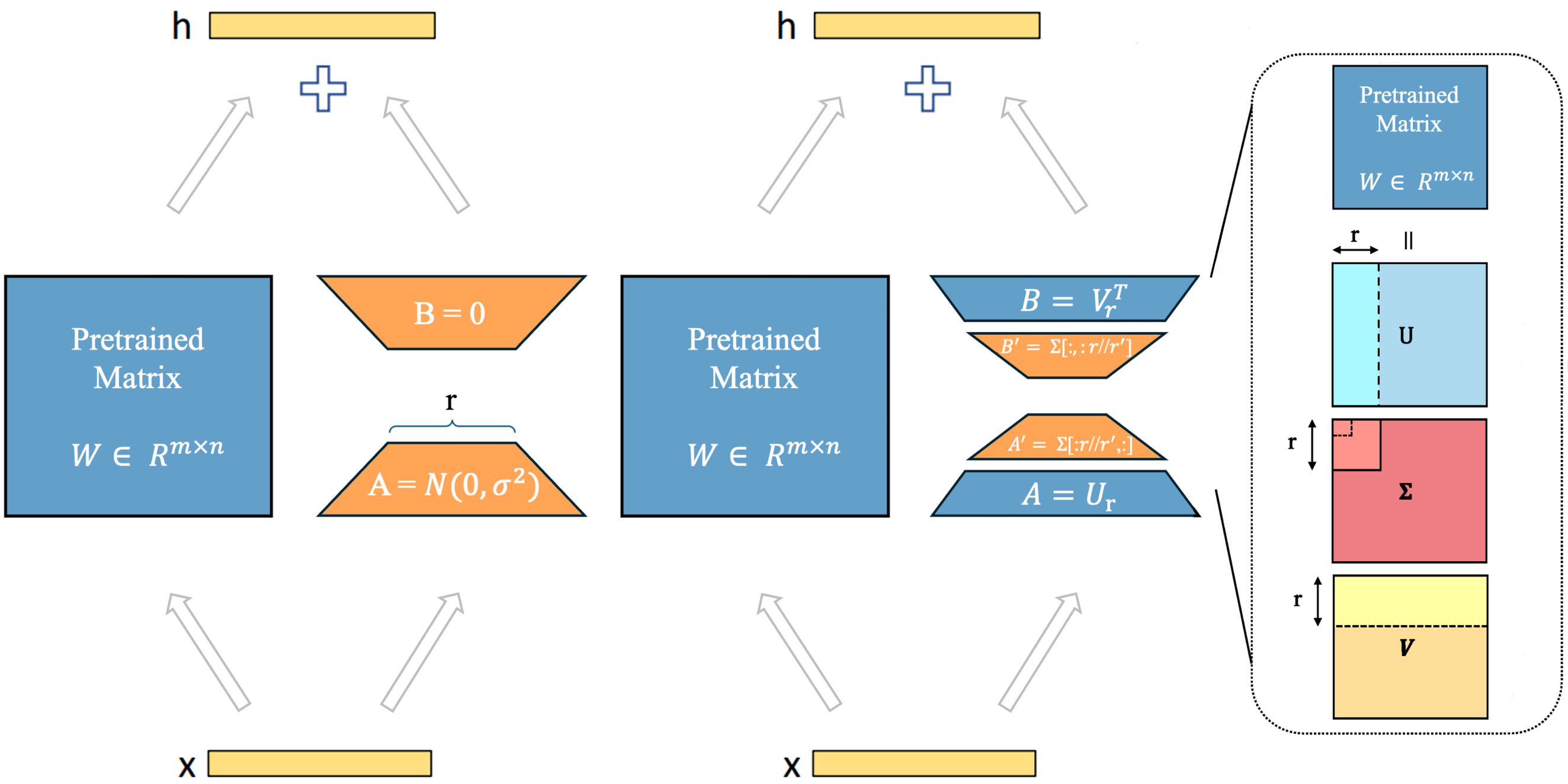}
    \caption{In the comparison between training with LoRA and NoRA, the blue modules represent the parts where parameters are frozen during training, while the orange modules indicate the components that need to be updated. Here, $r$ denotes the outer rank, and $r'$ denotes the inner rank.
    }
    \label{fig:image1}
\end{figure*}

LoRA utilizes low-rank matrices to achieve efficient and scalable adaptation to specific downstream tasks~\cite{houlsby2019parameter,he2021towards}, allowing the model to focus on essential parameters. This approach not only preserves the model's pre-trained knowledge but also facilitates its specialization with minimal computational overhead. The core hypothesis of LoRA is that the adaptations necessary to tailor a pre-trained large language model (LLM) to a specific task or domain are inherently low-dimensional, which can be effectively achieved through a low-rank decomposition of the weight matrices, minimizing computational overhead while preserving the model's pre-trained knowledge~\cite{mao2024survey}.

Despite LoRA's demonstrated utility, various LoRA variants proposed by recent researchers~\cite{liu2024aflora,tian2024hydralora}, each have their strengths and limitations. For example, while LoRA-FA~\cite{zhang2023lora} reduces activation memory demands by freezing part of the weights, it still suffers from the limitation of a fixed rank; VeRA~\cite{liu2023vera} enhances model scalability but remains sensitive to the hidden dimensions of the model; FLoRA~\cite{hao2024flora} and LoRA-XS~\cite{balazy2024lora} have made improvements in real-time performance and memory efficiency, yet they still do not fully address the complexity specific to tasks. 

These limitations prompted us to explore a new Nested Low-Rank Adaptation structure called NoRA, which enhances parameter efficiency and task adaptability through a dual-layer nested design combined with SVD. In the NoRA structure, the outer layer provides a stable low-rank foundation by keeping its weights fixed, while the inner layer introduces fine-grained adjustments tailored to different tasks. This innovative dual-layer nested structure effectively leverages the knowledge within the original matrix and further enhances model adaptability through precise optimization of certain parameters. NoRA not only retains the parameter efficiency advantages of LoRA but also improves the accuracy of low-rank matrix approximation through the decomposition properties of SVD, thereby enhancing the model's generalization capability and adaptability across different tasks. Moreover, the hierarchical separation between the outer and inner LoRA layers provides finer optimization control, enabling the model to more flexibly adjust the contribution of each layer's parameters, thus enhancing adaptability and flexibility.

We conducted experiments on multiple downstream tasks, including fine-tuning the LLaMA~\cite{touvron2023llama} model for commonsense reasoning, few-shot tuning on the CLIP~\cite{radford2021learning} model, and subject-driven generation on the Stable Diffusion XL~\cite{podell2023sdxl} model. NoRA effectively reduced the required parameters to as low as 10.2M for LLaMA3 8B while enhancing performance, achieving an average score of 85.0\%, surpassing LoRA's 82.8\%. Additionally, in visual few-shot tasks using ViT-B/16, NoRA achieved the highest average accuracies of 81.8\% (4 shots) and 85.4\% (16 shots), demonstrating its superior efficiency and effectiveness over existing methods. We summarize our contribution as below:
\begin{itemize}[leftmargin=2em]
    \item We propose NoRA, which uses a dual-layer nested structure and SVD. The outer LoRA provides a stable low-rank foundation, while the inner LoRA allows for precise task adjustments.
    \item NoRA leverages the original matrix's knowledge through its nested structure and SVD integration, enhancing low-rank matrix approximation accuracy. This design improves the model's adaptability across diverse tasks, while the hierarchical separation between layers allows for finer control over optimization.
    \item Our extensive experimental validation on multiple downstream tasks has shown that NoRA outperforms traditional LoRA and other recent variants such as VeRA and FLoRA.
\end{itemize}

\section{Related Work}

\noindent\textbf{Parameter-Efficient Fine-Tuning.} Parameter Efficient Fine-Tuning~\cite{pfeiffer2020adapterfusion, li2021prefix,zaken2021bitfit, liu2023pre, hu2021lora, hu2023llm} (PEFT) is a practical solution for adapting large pre-trained models to downstream tasks~\cite{zaken2021bitfit}. As the number of parameters in large-scale language models continues to grow, traditional fine-tuning methods face significant challenges, such as high computational resource demands and training costs~\cite{lv2023full}. PEFT optimizes the parameter adjustment process, effectively reducing the number of additional parameters introduced and the computational resources required for specific tasks or domains. The core idea is to enhance the model's task adaptability by streamlining parameter updates and introducing auxiliary modules while maintaining the structure and performance of the pre-trained model, without the need for complete retraining. This approach is particularly suitable for large-scale language models, as it reduces the computational burden and significantly improves the model's applicability across diverse downstream tasks. As a result, PEFT has become the mainstream trend for fine-tuning large pre-trained models, greatly promoting their adoption and application across different scenarios. Early PEFT research primarily focused on selective update strategies, such as Bias-Free Fine-Tuning~\cite{zaken2021bitfit} and Partial Network Training~\cite{gururangan2020don}, which achieve task-specific model fine-tuning by modifying only the most critical parameters of the pre-trained model. As research progressed, adapter modules became another direction in PEFT development, embedding additional modules, such as Prompt Tuning and Prefix Tuning, into the pre-trained architecture to further enhance the model's adaptability to specific tasks. Delta-weight techniques represent the latest innovation in PEFT, with LoRA and OFT~\cite{liu2023parameter} being notable examples. These strategies involve fine-tuning pre-trained parameters using trainable delta weights, providing fine-grained model adjustments that improve task-specific performance while minimizing computational overhead.

\noindent\textbf{Low-rank Adaptation.} 
LoRA has proven to be an efficient fine-tuning strategy in various task scenarios, leveraging low-rank decomposition to enhance model adaptation while minimizing computational overhead. However, LoRA's fixed rank limitation can restrict its flexibility in handling diverse tasks.  To address these limitations, researchers have proposed various LoRA variants, each showcasing unique strengths in enhancing model adaptability and performance. For instance, LoRA-FA~\cite{zhang2023lora} reduces activation memory demands by freezing part of the weights, though it remains limited by the fixed rank; VeRA~\cite{liu2023vera} excels in enhancing model scalability but is still sensitive to the model's hidden dimensions; FLoRA~\cite{hao2024flora} and LoRA-XS~\cite{zhang2023lora} improve real-time performance and memory efficiency, yet they do not fully address the complexity specific to certain tasks. Other variants, such as VB-LoRA~\cite{li2024vb} and Tied-LoRA~\cite{renduchintala2023tied}, introduce new adaptation mechanisms that enhance model adaptability but also bring additional computational overhead and complexity, posing challenges in practical applications.  Moreover, as research has progressed, techniques such as VeRA and DoRA~\cite{liu2024dora} have further optimized the LoRA method, making innovative improvements in parameter efficiency, as well as in the distribution and structure of update matrices. Advanced methods like AdaLoRA~\cite{adalora} and PiSSA~\cite{meng2024pissa} push the limits of parameter update efficiency by selectively adjusting matrix ranks and distributions, significantly improving the applicability of large-scale pre-trained models in complex tasks. However, despite the significant advancements achieved by these LoRA variants, they still face certain limitations and challenges in practical applications, particularly in meeting diverse task requirements and optimizing computational resource usage, which necessitate further research and solutions.

\section{Methodology: Nested Low-Rank Adaptation}
In this paper, we present NoRA, a novel approach to parameter-efficient fine-tuning that builds upon traditional LoRA techniques. By incorporating a dual-layer nested structure with SVD, NoRA effectively harnesses the knowledge of the original matrix, further reducing fine-tuning parameters while enhancing the model's adaptability and optimization control.

\subsection{Review of Low-Rank Adaptation}
LoRA is a parameter-efficient method for fine-tuning large-scale pre-trained models. It achieves fine-tuning of the original weights through the introduction of low-rank matrix updates, without altering the stability and performance of the pre-trained models. In LoRA, the weights for each layer are updated using the following formula:
\begin{equation}
h = Wx + \Delta Wx = Wx + BAx ,
\end{equation}
where \(\Delta W \in \mathbb{R}^{m \times n}\) is the low-rank weight update, and \(A \in \mathbb{R}^{r \times n}\) and \(B \in \mathbb{R}^{m \times r}\) are the low-rank matrices with \(r \ll \min(m, n)\). During training, \(W\) is kept frozen, and \(A\) and \(B\) are the trainable parameters.

\subsection{ NoRA Structure and Initialization}
NoRA employs a dual-layer nested structure, where each layer is initialized using the SVD~\cite{golub2013matrix} of the pre-trained weights to enhance the model's learning capabilities. Specifically, NoRA first decomposes the original weights using SVD, then initializes the matrix \(A\) with \(U\) and the matrix \(B\) with \(V^T\). The inner LoRA weights are initialized using the intermediate diagonal matrix, thus optimizing the weight updates \(\Delta W\).

\begin{itemize}[leftmargin=2em]
    \item \textbf{Outer LoRA Layer}: This layer is initialized using the SVD results \(U \Sigma V^T\) of the pre-trained weights \(W\). The parameters of this outer LoRA layer are frozen during training. Freezing these parameters helps maintain stability and preserve the key features of the pre-trained model, while still allowing for precise adjustments through the inner LoRA layer.
    \item \textbf{Inner LoRA Layer}: Initialized with the diagonal matrix \(\Sigma\) from the SVD~\cite{golub2013matrix} of the \( U \Sigma V^T\). Initializing with \(\Sigma\) allows this layer to focus on subtle perturbations in the weight space, enabling finer adjustments without altering the core weights preserved by the outer LoRA layer. This approach ensures that updates are focused on refining and enhancing the model's ability to adapt to new tasks, leveraging minor adjustments that have a targeted impact on the model's performance.
\end{itemize}
The final weight update formula is:
\begin{equation}
h = Wx + \Delta Wx = Wx + BAx + B'A'x ,
\end{equation}

\begin{wrapfigure}{t}{0.525\textwidth}
\vspace{-3ex}
\begin{minipage}[t]{0.525\textwidth}
\begin{algorithm}[H]
\caption{PyTorch code for NoRA}
\label{alg:wanda-pytorch}
\vspace{-1.ex}
\small
\begin{lstlisting}[style=Pytorch,escapeinside={(@}{@)}]
# r_out: rank of the outer LoRA layer.
# r_in: rank of the inner LoRA layer.
# A^ and B^ are the inner LoRA weights.

def init nora param(weight, r_out, r_in):
    U, S, V = torch.svd(weight)

    A = U[:, :r_out]
    B = V.T[:r_out, :]

    S = torch.diag(S)
    S_diag = torch.sqrt(S[:r_out, :r_out])

    A^ = S_diag[:, :r_out // r_in] 
    B^ = S_diag[:r_out // r_in, :]
\end{lstlisting}
\vspace{-1.ex}
\end{algorithm}
\end{minipage}
\vspace{-1ex}
\end{wrapfigure}

where \(A\) and \(B\) are set using the truncated SVD of the original weight matrix \(W\), with the outer LoRA weights remaining frozen. The matrices \(A\) and \(B\) are initialized as \(A = U_r\) and \(B = V_r^T\) where \(U_r\) and \(V_r\) are the left and right singular vectors corresponding to the top \(r\) singular values, and these vectors are obtained from the decomposition: $W = U \Sigma V^T$ with \(U \in \mathbb{R}^{m \times m}\), \(\Sigma \in \mathbb{R}^{m \times n}\), and \(V \in \mathbb{R}^{n \times n}\). For the inner LoRA layer, the matrices \(B'\) and \(A'\) are initialized by taking the square roots of selected components from the top \(r_{in}\) singular values in the diagonal matrix \(\Sigma_r\), where \(\Sigma_r \in \mathbb{R}^{r \times r}\). Specifically, \(B'\) is initialized using the square roots of the first \(r_{in}\) columns of \(\Sigma_r\), and \(A'\) is initialized using the square roots of the first \(r_{in}\) rows of \(\Sigma_r\). This initialization approach introduces another layer of adaptation within the NoRA framework by decomposing the smaller matrix \(W_r\) through an additional LoRA process.

\subsection{Parameter Efficiency of NoRA}
We make the following observation on the parameter efficiency of NoRA compared to LoRA and VeRA methods.
Observation: NoRA demonstrates superior parameter efficiency compared to both LoRA and VeRA.
For simplicity, let’s consider a transformer model with \(L\) finetuned layers, each consisting of \(q\) number of \(W \in \mathbb{R}^{n \times n}\) matrices.

For LoRA, the number of trainable parameters is given by:
\begin{equation}
P_{\text{LoRA}} = L \times q \times r \times 2n .
\end{equation}

For NoRA, the number of trainable parameters is given by:
\begin{equation}
P_{\text{NoRA}} = L \times q \times r_{out} \times r_{in} \times 2 .
\end{equation}
To compare the parameter efficiency, we compute the ratios of the number of trainable parameters between the methods. The ratio of trainable parameters for LoRA to NoRA is:
\begin{equation}
\frac{P_{\text{LoRA}}}{P_{\text{NoRA}}} = \frac{L \times q \times r \times 2n}{L \times q \times r_{out} \times r_{in} \times 2} = \frac{r \times n}{r_{out} \times r_{in}}.
\end{equation}
Since the inner rank ($r_{\text{in}}$) in the NoRA model is generally set to be the same as the rank ($r$) used in the LoRA model, NoRA demonstrates more pronounced advantages over LoRA under certain conditions. Particularly when the outer rank ($r_{\text{out}}$) is less than the maximum allowable rank ($r_{\text{max}}$), the NoRA model exhibits significantly enhanced parameter optimization. Specifically, the ratio of parameter efficiency between LoRA and NoRA can be approximated as follows:
\begin{equation}
\frac{P_{\text{LoRA}}}{P_{\text{NoRA}}} \approx \frac{n}{r_{out}}.
\end{equation}

\section{Experiment}
\subsection{Fine-tuning of Large Language Models}
\noindent\textbf{Implementation Details.}
We employed the LLaMA3 8B~\cite{touvron2023llama} and LLaMA 7B models, each fine-tuned using a variety of parameter-efficient methods to enhance their commonsense reasoning capabilities. We utilized the Commonsense170K dataset~\cite{li2016commonsense} for targeted fine-tuning, which is designed to enhance the models' understanding of commonsense knowledge across different contexts. The main objective was to assess the effectiveness of each fine-tuning approach in improving the model's performance on a range of commonsense reasoning tasks. Post-fine-tuning, model performance was evaluated using a suite of eight benchmark tests focused on commonsense reasoning, including ARC-e, OBQA, SIQA, and more.

\noindent\textbf{Comparison Results.}
The results from the experimental evaluations are detailed in Table~\ref{tab2}. The fine-tuning methods demonstrated varying levels of success in enhancing the reasoning capabilities of the LLaMA models.

The LLaMA 7B model, when fine-tuned with the NoRA method, exhibited the highest average score of 75.8 across most tasks, demonstrating its superior ability to generalize across different question sets. Remarkably, it achieved top scores in HellaSwag (80.6), WinoGrande (79.6), and ARC-e (80.5), underscoring its robust understanding and reasoning capabilities. The NoRA method not only showed the highest scores but also significantly reduced parameter utilization, emphasizing its efficiency and scalability as a fine-tuning strategy. In contrast, other methods like LoKr and AdaLoRA, while effective in certain areas, required more parameters or resulted in higher computational costs. Notably, the NoRA method outperformed others in terms of parameter efficiency and computational overhead, making it a promising approach for practical applications where resource constraints are a concern.

These findings highlight the potential of NoRA as a scalable and effective fine-tuning strategy that not only achieves high performance across diverse datasets but also maintains a low parameter footprint, enhancing the practical usability of large pre-trained models in varied commonsense reasoning applications.

\subsection{Fine-tuning of Vision-Language Models}
\noindent\textbf{Implementation Details.}
The performance of various adaptation techniques on the Vision Transformer~\cite{dosovitskiy2020image} (ViT-B/16) was evaluated across three distinct datasets: Food, Pets, and DTD. Each dataset was chosen to test the robustness and adaptability of these methods across different visual domains. The primary performance metric used was the Top-1 accuracy, which was computed as an average over three random seeds to mitigate randomness and ensure reliability in the results. The experiments were conducted under two shot settings, 4 and 16 shots, to determine the effectiveness of each adaptation technique under limited data conditions.

The adaptation methods tested included various enhancements of existing techniques such as CoOp~\cite{zhou2022learning},~\cite{zhou2022learning} PLOT++~\cite{chen2022plot}, MaPLe~\cite{khattak2023maple}, and several variants of LoRA, such as CLIP-LoRA~\cite{zanella2024low}, AdaLoRA~\cite{adalora}, DyLoRA~\cite{valipour2022dylora}, and others. This comprehensive approach allowed for a detailed assessment of each method’s ability to improve model performance under constrained training scenarios.

\begin{table}[t]
\centering
\caption{ Averaged accuracies (\%) for 8 zero-shot tasks. Param denotes the number of trained parameters, Time for the training time on H800 GPU, and Mem for the GPU Memory usage.}
\resizebox{140mm}{!}{
\begin{tabular}{lcccccccccccc}
\toprule
 Method & Param & Time & Mem & BoolQ & PIQA & SIQA & HellaSwag & WinoGrande & ARC-e & ARC-c & OBQA & Avg. \\
         \multicolumn{12}{c}{\textbf{\textit{Fine-tuning on LLaMA-1 7B}}}\\
LoRA$_{r=16}$~\cite{lora} & 8.4M & 5.7h & 21G & 68.9 & 80.7 & 77.4 & 78.1 & 78.8 & 77.8 &  61.3 &  74.8 & 74.4 \\
 LoRA$_{r=32}$~\cite{lora} & 16.8M & 6.5h & 27G & 68.5 & 81.0 & 77.4 & 77.1 & 79.0 & 77.8 & 63.3 & 77.9 & 75.3 \\
\textbf{NoRA}& 8.2M & 5.0h & 19G & 68.1 & 80.3 & 76.8 & 80.6 & 79.6 & 80.5 & 62.6 & 77.8 & 75.8 \\
          \multicolumn{12}{c}{\textbf{\textit{Fine-tuning on LLaMA-3 8B}}}\\
LoRA~\cite{lora} & 28.3M & 8.0h & 29G & 72.3 & 86.7 & 79.3 & 93.5 & 84.8 & 87.7 & 75.7 & 82.8 & 82.8 \\
LoKr~\cite{yeh2023navigating} & 0.9M & 26.3h & 66G & 65.1 & 81.6 & 78.7 & 92.0 & 82.1 & 89.2 & 76.7 & 80.9 & 80.9 \\
 AdaLoRA~\cite{adalora} & 28.3M & 12.5h & 58G & 75.1 & 86.4 & 76.7 & 75.4 & 83.3 & 90.4 & 79.1 & 81.4 & 81.4 \\
\textbf{NoRA} & 7.2M & 6.2h & 28G & 73.3 & 86.4 & 79.1 & 94.1 & 84.3 & 88.2 & 77.5 & 85.0 & 83.1\\
\bottomrule
\end{tabular}
}
\label{tab2}
\end{table}

\begin{table}[t]
    \caption{Detailed results for 3 datasets with the ViT-B/16 as visual backbone. Top-1 accuracy averaged over 3 random seeds is reported.
Highest value is highlighted in bold, and the second highest is underlined.}
    \vspace{2pt}
    \label{tab3}
    \centering
    \resizebox{\textwidth}{!}{
    \begin{tabular}{cccccc}
        \toprule
        \multicolumn{5}{c}{Shots 4}\\ \midrule
        Method & Food & Pets & DTD  & Average\\ 
        \hline
        CoOp~\cite{zhou2022learning} (4) & 83.5 & 92.3 & 58.5 & 78.1 \\
        CoOp~\cite{zhou2022learning} (16) & 84.5 & 92.5 & 59.5 & 78.8\\
        CoCoOp~\cite{zhou2022conditional} & 86.3 &  92.7 & 55.7 & 78.2 \\
       TIP-Adapter-F~\cite{zhang2022tip}  &  86.5 & 91.9 & 59.8 & 79.4\\
       CLIP-Adapter~\cite{gao2024clip} & 86.5 &  90.8 & 46.1 & 74.5\\
        PLOT++~\cite{chen2022plot} & 86.5 & 92.6 & 62.4 & 80.5\\
        KgCoOp~\cite{yao2023visual} & 86.9 & 92.6 & 58.7 & 79.4\\
        TaskRes~\cite{yu2023task}&  86.0 &  91.9 & 60.1 & 79.3\\
        MaPLe~\cite{khattak2023maple} &  86.7 & \textbf{93.3} & 59.0 & 79.7\\
        ProGrad~\cite{zhu2023prompt}&  85.4 &  92.1 & 59.7 & 79.1\\
        CLIP-LoRA~\cite{zanella2024low}& 82.7 & 91.0 & 63.8 & 79.2 \\
        LoRA+~\cite{hayou2024lora+}& 84.4 & 92.8 & 64.1 & 81.4 \\
        AdaLoRA~\cite{adalora}& 85.6 & 92.8 & \textbf{66.2} & \underline{81.5} \\
        DyLoRA~\cite{valipour2022dylora}& \underline{87.0} & 92.4 & 64.9 & 81.4 \\
        LoRA-FA~\cite{zhang2023lora}& 86.7 & 93.0 & 64.4 & 81.4 \\
        VeRA~\cite{liu2023vera}& 84.5 & 92.5 & 65.1 & 80.7 \\
	  \textbf{NoRA} & \textbf{87.1} & \underline{93.1} & \underline{65.2} & \textbf{81.8} \\
        \bottomrule
    \end{tabular}
    \quad
    \begin{tabular}{ccccc}
        \toprule
        \multicolumn{5}{c}{Shots 16}\\ \midrule
        Method & Food & Pets & DTD &  Average \\ 
        \hline
        CoOp~\cite{zhou2022learning} (4) & 85.1 & 92.4 & 81.2  & 86.2\\
        CoOp~\cite{zhou2022learning} (16) & 84.2 & 92.0 & 69.7 & 81.9\\
        CoCoOp~\cite{zhou2022conditional} & 87.4 & 93.4 & 63.7 & 81.5\\
       TIP-Adapter-F~\cite{zhang2022tip}  & 86.8 & 92.6 & 70.8 & 83.4\\
       CLIP-Adapter~\cite{zhang2022tip} & 87.1 & 92.3 & 59.4 & 79.6\\
        PLOT++~\cite{chen2022plot} & 87.1 & 93.6 & 71.4 & 84.0\\
        KgCoOp~\cite{yao2023visual} & 87.2 & 93.2 & 68.7 & 83.0\\
        TaskRes~\cite{yu2023task}& 86.9 & 92.4  & 71.5 & 83.6\\
        MaPLe~\cite{khattak2023maple} & 87.4 & 93.2  & 68.4 & 83.0\\
        ProGrad~\cite{zhu2023prompt}& 85.8 & 92.8 & 68.8 & 82.5\\
        CLIP-LoRA~\cite{zanella2024low}& 84.2 & 92.4  & 72.0 & 82.9\\
        LoRA+~\cite{hayou2024lora+}& 85.1 & 93.6 & 72.1 & 83.6 \\
        AdaLoRA~\cite{adalora}& 85.9 & 93.7 & \underline{72.8} &  84.1 \\
        DyLoRA~\cite{valipour2022dylora}& \underline{87.6} & 93.0 & 72.7 & \underline{84.4} \\
        LoRA-FA~\cite{zhang2023lora}& 87.4 & \underline{93.9} & 71.9 & \underline{84.4} \\
        VeRA~\cite{liu2023vera}& 86.2 & 92.2 & 72.2  & 83.5 \\
	  \textbf{NoRA} & \textbf{87.8} & \textbf{94.1} & \textbf{74.3} & \textbf{85.4}\\
        \bottomrule
    \end{tabular}
    }
\vspace{-2mm}
\end{table}

\noindent\textbf{Comparison Results.}
The detailed results are presented in Table~\ref{tab3}, which includes the Top-1 accuracy for each method across the three datasets under both 4-shot and 16-shot settings. Notably, the NoRA model consistently outperformed other adaptation methods in both settings, highlighting its superior adaptability and efficiency. In the 4-shot setting, NoRA achieved an average Top-1 accuracy rate of 81.8, slightly surpassing DyLoRA, the second-best method, by 0.2\%. In the more demanding 16-shot setting, NoRA demonstrated even greater efficacy, achieving an average Top-1 accuracy of 85.4 and surpassing DyLoRA’s score of 85.0.

NoRA showed exceptional robustness across the various visual domains tested. Specifically, in the 16-shot setting, NoRA achieved the best results in all individual datasets, with scores of 87.8 for Food, 94.1 for Pets, and 74.3 for DTD. These scores underline NoRA’s capability to adapt to and perform well across diverse categories, ensuring high reliability for practical applications in fields requiring visual recognition.

Overall, the experimental outcomes underscore the efficacy of NoRA as a scalable and effective fine-tuning strategy, achieving high performance across diverse datasets and maintaining a low parameter footprint, which is crucial for practical deployments where computational efficiency is essential. These results confirm the importance of selecting appropriate adaptation techniques to enhance the performance of vision transformers in limited data environments.

\begin{figure}[t] 
    \centering 
    \includegraphics[width=1\textwidth]{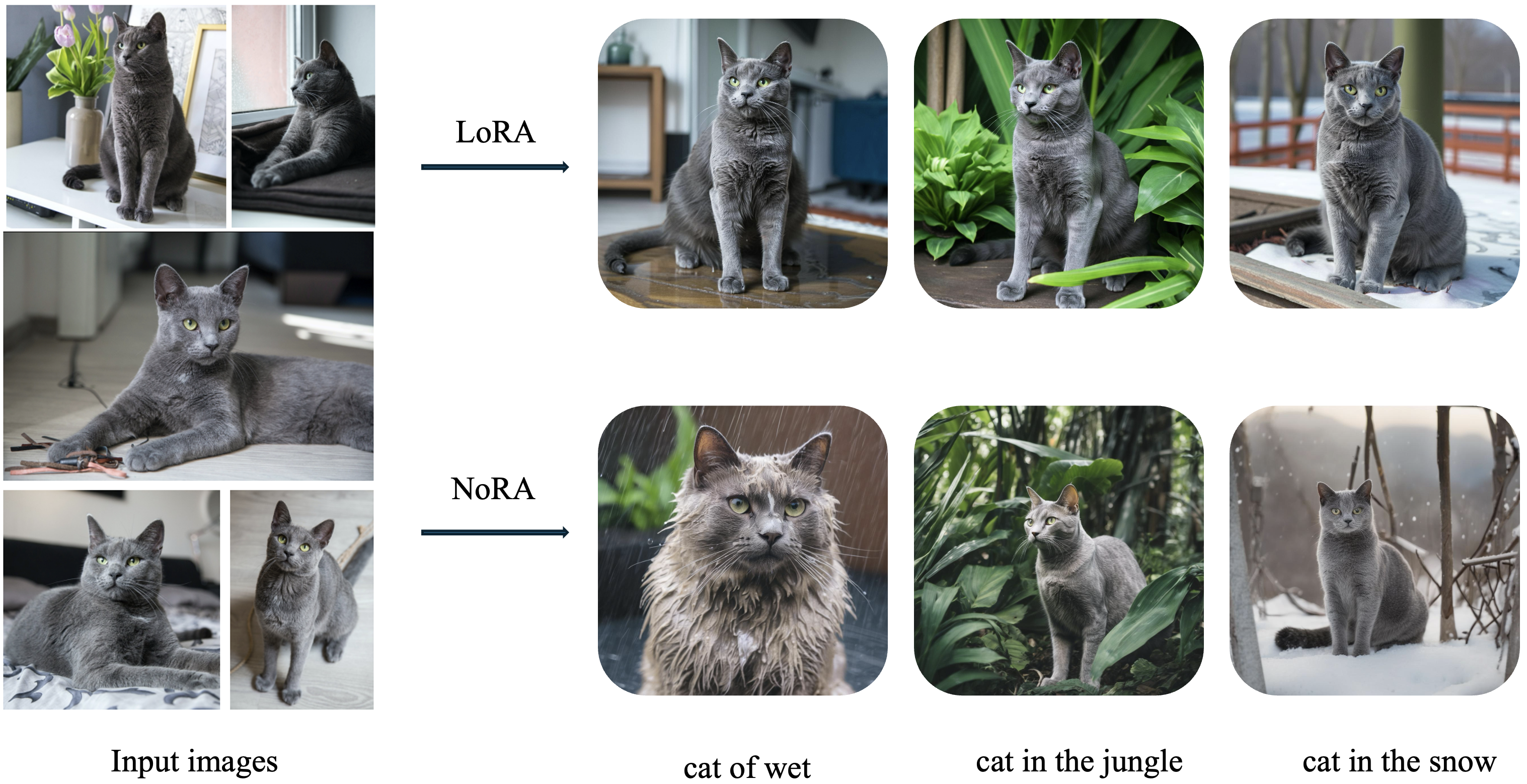} 
    \caption{Comparison of generated images from LoRA and NoRA on the subject-driven generation task.} 
    \label{fig:image2} 
\end{figure}
\subsection{Subject-driven Generation}
\noindent\textbf{Implementation Details.}
We explored theme-based image generation by utilizing advanced text-to-image diffusion models, specifically employing a pre-trained text-to-image model. This model was fine-tuned using a combination of images and specific textual prompts (e.g., "[V] photo of a cat"). The fine-tuning involved sophisticated adaptation techniques, namely the LoRA and NoRA methods, to enhance the model's capability to generate images that align closely with given themes.

The model used in our experiments, referred to as SDXL5, was fine-tuned on a 32G V100S GPU. The parameters adjusted during fine-tuning included setting the learning rate to 1e-4 and the batch size to 4. The NoRA method involved initializing the internal and external layers of LoRA with a differential diagonal matrix. The training was conducted over 500 steps, taking approximately 24 minutes to complete.

\noindent\textbf{Comparison Results.}
The results of the image generation process are detailed in Figure~\ref{fig:image2}. During the image generation phase, we performed 50 inference steps for each textual prompt. Our findings indicate that the NoRA method outperformed the traditional LoRA method in several key aspects. Notably, the NoRA method was more effective in capturing complex themes and details within the images. The images generated using the NoRA method showed a higher degree of visual alignment with the specified textual prompts. They better reflected specific environmental contexts, such as forests and snowy scenes, and captured intricate object details, like wet fur.

These outcomes underscore the efficacy of the NoRA method in enhancing thematic consistency and visual expressiveness. The enhanced model demonstrates significant improvements in generating thematic images that are closely aligned with the nuances of the input prompts, offering promising applications in fields requiring detailed and context-specific image generation. This experiment sets a foundation for further exploration into fine-tuning techniques that can more adeptly handle complex thematic prompts in image generation tasks.

\section{Conclusion}
In this paper we present  introduces NoRA, a new parameter-efficient fine-tuning method for large models. NoRA employs a dual-layer nested structure combined with SVD to extract latent information from the original matrix while reducing parameter count. The method retains LoRA's parameter efficiency while improving low-rank matrix approximation accuracy through SVD. NoRA's hierarchical structure allows for finer optimization control, enhancing model adaptability across tasks. Experimental results demonstrate NoRA's effectiveness across various tasks, including commonsense reasoning in LLMs, VLMs and subject-driven generation. The significance of NoRA lies in its ability to enhance model adaptability and performance while maintaining excellent parameter efficiency. Our approach addresses the growing need for efficient fine-tuning methods in the era of large models.

Limitations of the study include potential computational overhead due to SVD calculations and the need for further investigation into NoRA's scalability to even larger models. Future work could explore combining NoRA with AutoML~\cite{dong2023emq,Dong2023diswot,Pruner-Zero,wei2024auto,li2024attnzero,li2024Autogas} and distillation techniques~\cite{lishadow,li2024kd,li2022self,li2022norm,li2024detkds}, applying it to multimodal models, and investigating its impact on model interpretability and robustness.


\bibliography{main}

\begin{thebibliography}{10}

\bibitem{balazy2024lora}
Klaudia Ba{\l}azy, Mohammadreza Banaei, Karl Aberer, and Jacek Tabor.
\newblock Lora-xs: Low-rank adaptation with extremely small number of parameters.
\newblock {\em arXiv preprint arXiv:2405.17604}, 2024.

\bibitem{chen2022plot}
Guangyi Chen, Weiran Yao, Xiangchen Song, Xinyue Li, Yongming Rao, and Kun Zhang.
\newblock Plot: Prompt learning with optimal transport for vision-language models.
\newblock {\em arXiv preprint arXiv:2210.01253}, 2022.

\bibitem{ding2023parameter}
Ning Ding, Yujia Qin, Guang Yang, Fuchao Wei, Zonghan Yang, Yusheng Su, Shengding Hu, Yulin Chen, Chi-Min Chan, Weize Chen, et~al.
\newblock Parameter-efficient fine-tuning of large-scale pre-trained language models.
\newblock {\em Nature Machine Intelligence}, 5(3):220--235, 2023.

\bibitem{Pruner-Zero}
Peijie Dong, Lujun Li, Zhenheng Tang, Xiang Liu, Xinglin Pan, Qiang Wang, and Xiaowen Chu.
\newblock Pruner-zero: Evolving symbolic pruning metric from scratch for large language models.
\newblock In {\em ICML}, 2024.

\bibitem{Dong2023diswot}
Peijie Dong, Lujun Li, and Zimian Wei.
\newblock Diswot: Student architecture search for distillation without training.
\newblock In {\em CVPR}, 2023.

\bibitem{dong2023emq}
Peijie Dong, Lujun Li, Zimian Wei, Xin Niu, Zhiliang Tian, and Hengyue Pan.
\newblock Emq: Evolving training-free proxies for automated mixed precision quantization.
\newblock In {\em ICCV}, 2023.

\bibitem{dosovitskiy2020image}
Alexey Dosovitskiy, Lucas Beyer, Alexander Kolesnikov, Dirk Weissenborn, Xiaohua Zhai, Thomas Unterthiner, Mostafa Dehghani, Matthias Minderer, Georg Heigold, Sylvain Gelly, et~al.
\newblock An image is worth 16x16 words: Transformers for image recognition at scale.
\newblock {\em arXiv preprint arXiv:2010.11929}, 2020.

\bibitem{gao2024clip}
Peng Gao, Shijie Geng, Renrui Zhang, Teli Ma, Rongyao Fang, Yongfeng Zhang, Hongsheng Li, and Yu~Qiao.
\newblock Clip-adapter: Better vision-language models with feature adapters.
\newblock {\em International Journal of Computer Vision}, 132(2):581--595, 2024.

\bibitem{golub2013matrix}
Gene~H Golub and Charles~F Van~Loan.
\newblock {\em Matrix computations}.
\newblock JHU press, 2013.

\bibitem{gururangan2020don}
Suchin Gururangan, Ana Marasovi{\'c}, Swabha Swayamdipta, Kyle Lo, Iz~Beltagy, Doug Downey, and Noah~A Smith.
\newblock Don't stop pretraining: Adapt language models to domains and tasks.
\newblock {\em arXiv preprint arXiv:2004.10964}, 2020.

\bibitem{han2024parameter}
Zeyu Han, Chao Gao, Jinyang Liu, Sai~Qian Zhang, et~al.
\newblock Parameter-efficient fine-tuning for large models: A comprehensive survey.
\newblock {\em arXiv preprint arXiv:2403.14608}, 2024.

\bibitem{hao2024flora}
Yongchang Hao, Yanshuai Cao, and Lili Mou.
\newblock Flora: Low-rank adapters are secretly gradient compressors.
\newblock {\em arXiv preprint arXiv:2402.03293}, 2024.

\bibitem{hayou2024lora+}
Soufiane Hayou, Nikhil Ghosh, and Bin Yu.
\newblock Lora+: Efficient low rank adaptation of large models.
\newblock {\em arXiv preprint arXiv:2402.12354}, 2024.

\bibitem{he2021towards}
Junxian He, Chunting Zhou, Xuezhe Ma, Taylor Berg-Kirkpatrick, and Graham Neubig.
\newblock Towards a unified view of parameter-efficient transfer learning.
\newblock {\em arXiv preprint arXiv:2110.04366}, 2021.

\bibitem{houlsby2019parameter}
Neil Houlsby, Andrei Giurgiu, Stanislaw Jastrzebski, Bruna Morrone, Quentin De~Laroussilhe, Andrea Gesmundo, Mona Attariyan, and Sylvain Gelly.
\newblock Parameter-efficient transfer learning for nlp.
\newblock In {\em International conference on machine learning}, pages 2790--2799. PMLR, 2019.

\bibitem{hu2021lora}
Edward~J Hu, Yelong Shen, Phillip Wallis, Zeyuan Allen-Zhu, Yuanzhi Li, Shean Wang, Lu~Wang, and Weizhu Chen.
\newblock Lora: Low-rank adaptation of large language models.
\newblock {\em arXiv preprint arXiv:2106.09685}, 2021.

\bibitem{lora}
Edward~J Hu, Yelong Shen, Phillip Wallis, Zeyuan Allen-Zhu, Yuanzhi Li, Shean Wang, Lu~Wang, and Weizhu Chen.
\newblock Lora: Low-rank adaptation of large language models.
\newblock {\em arXiv preprint arXiv:2106.09685}, 2021.

\bibitem{hu2023llm}
Zhiqiang Hu, Lei Wang, Yihuai Lan, Wanyu Xu, Ee-Peng Lim, Lidong Bing, Xing Xu, Soujanya Poria, and Roy Ka-Wei Lee.
\newblock Llm-adapters: An adapter family for parameter-efficient fine-tuning of large language models.
\newblock {\em arXiv preprint arXiv:2304.01933}, 2023.

\bibitem{khattak2023maple}
Muhammad~Uzair Khattak, Hanoona Rasheed, Muhammad Maaz, Salman Khan, and Fahad~Shahbaz Khan.
\newblock Maple: Multi-modal prompt learning.
\newblock In {\em Proceedings of the IEEE/CVF Conference on Computer Vision and Pattern Recognition}, pages 19113--19122, 2023.

\bibitem{li2022self}
Lujun Li.
\newblock Self-regulated feature learning via teacher-free feature distillation.
\newblock In {\em ECCV}, 2022.

\bibitem{li2024detkds}
Lujun Li, Yufan Bao, Peijie Dong, Chuanguang Yang, Anggeng Li, Wenhan Luo, Qifeng Liu, Wei Xue, and Yike Guo.
\newblock Detkds: Knowledge distillation search for object detectors.
\newblock In {\em ICML}, 2024.

\bibitem{li2024kd}
Lujun Li, Peijie Dong, Anggeng Li, Zimian Wei, and Ya~Yang.
\newblock Kd-zero: Evolving knowledge distiller for any teacher-student pairs.
\newblock {\em NeuIPS}, 2024.

\bibitem{lishadow}
Lujun Li and Zhe Jin.
\newblock Shadow knowledge distillation: Bridging offline and online knowledge transfer.
\newblock In {\em NeuIPS}, 2022.

\bibitem{li2024Autogas}
Lujun Li, Haosen Sun, Shiwen Li, Peijie Dong, Wenhan Luo, Wei Xue, Qifeng Liu, and Yike. Guo.
\newblock Auto-gas: Automated proxy discovery for training-free generative architecture search.
\newblock In {\em ECCV}, 2024.

\bibitem{li2024attnzero}
Lujun Li, Zimian Wei, Peijie Dong, Wenhan Luo, Wei Xue, Qifeng Liu, and Yike. Guo.
\newblock Attnzero: Efficient attention discovery for vision transformers.
\newblock In {\em ECCV}, 2024.

\bibitem{li2016commonsense}
Xiang Li, Aynaz Taheri, Lifu Tu, and Kevin Gimpel.
\newblock Commonsense knowledge base completion.
\newblock In {\em Proceedings of the 54th Annual Meeting of the Association for Computational Linguistics (Volume 1: Long Papers)}, pages 1445--1455, 2016.

\bibitem{li2021prefix}
Xiang~Lisa Li and Percy Liang.
\newblock Prefix-tuning: Optimizing continuous prompts for generation.
\newblock {\em arXiv preprint arXiv:2101.00190}, 2021.

\bibitem{li2024vb}
Yang Li, Shaobo Han, and Shihao Ji.
\newblock Vb-lora: Extreme parameter efficient fine-tuning with vector banks.
\newblock {\em arXiv preprint arXiv:2405.15179}, 2024.

\bibitem{lin2024sdxl}
Shanchuan Lin, Anran Wang, and Xiao Yang.
\newblock Sdxl-lightning: Progressive adversarial diffusion distillation.
\newblock {\em arXiv preprint arXiv:2402.13929}, 2024.

\bibitem{liu2023vera}
Jiacheng Liu, Wenya Wang, Dianzhuo Wang, Noah~A Smith, Yejin Choi, and Hannaneh Hajishirzi.
\newblock Vera: A general-purpose plausibility estimation model for commonsense statements.
\newblock {\em arXiv preprint arXiv:2305.03695}, 2023.

\bibitem{liu2023pre}
Pengfei Liu, Weizhe Yuan, Jinlan Fu, Zhengbao Jiang, Hiroaki Hayashi, and Graham Neubig.
\newblock Pre-train, prompt, and predict: A systematic survey of prompting methods in natural language processing.
\newblock {\em ACM Computing Surveys}, 55(9):1--35, 2023.

\bibitem{liu2024dora}
Shih-Yang Liu, Chien-Yi Wang, Hongxu Yin, Pavlo Molchanov, Yu-Chiang~Frank Wang, Kwang-Ting Cheng, and Min-Hung Chen.
\newblock Dora: Weight-decomposed low-rank adaptation.
\newblock {\em arXiv preprint arXiv:2402.09353}, 2024.

\bibitem{dora}
Shih-Yang Liu, Chien-Yi Wang, Hongxu Yin, Pavlo Molchanov, Yu-Chiang~Frank Wang, Kwang-Ting Cheng, and Min-Hung Chen.
\newblock Dora: Weight-decomposed low-rank adaptation.
\newblock {\em arXiv preprint arXiv:2402.09353}, 2024.

\bibitem{liu2023parameter}
Weiyang Liu, Zeju Qiu, Yao Feng, Yuliang Xiu, Yuxuan Xue, Longhui Yu, Haiwen Feng, Zhen Liu, Juyeon Heo, Songyou Peng, et~al.
\newblock Parameter-efficient orthogonal finetuning via butterfly factorization.
\newblock {\em arXiv preprint arXiv:2311.06243}, 2023.

\bibitem{liu2024aflora}
Zeyu Liu, Souvik Kundu, Anni Li, Junrui Wan, Lianghao Jiang, and Peter~Anthony Beerel.
\newblock Aflora: Adaptive freezing of low rank adaptation in parameter efficient fine-tuning of large models.
\newblock {\em arXiv preprint arXiv:2403.13269}, 2024.

\bibitem{lv2023full}
Kai Lv, Yuqing Yang, Tengxiao Liu, Qinghui Gao, Qipeng Guo, and Xipeng Qiu.
\newblock Full parameter fine-tuning for large language models with limited resources.
\newblock {\em arXiv preprint arXiv:2306.09782}, 2023.

\bibitem{mao2024survey}
Yuren Mao, Yuhang Ge, Yijiang Fan, Wenyi Xu, Yu~Mi, Zhonghao Hu, and Yunjun Gao.
\newblock A survey on lora of large language models.
\newblock {\em arXiv preprint arXiv:2407.11046}, 2024.

\bibitem{meng2024pissa}
Fanxu Meng, Zhaohui Wang, and Muhan Zhang.
\newblock Pissa: Principal singular values and singular vectors adaptation of large language models.
\newblock {\em arXiv preprint arXiv:2404.02948}, 2024.

\bibitem{pfeiffer2020adapterfusion}
Jonas Pfeiffer, Aishwarya Kamath, Andreas R{\"u}ckl{\'e}, Kyunghyun Cho, and Iryna Gurevych.
\newblock Adapterfusion: Non-destructive task composition for transfer learning.
\newblock {\em arXiv preprint arXiv:2005.00247}, 2020.

\bibitem{podell2023sdxl}
Dustin Podell, Zion English, Kyle Lacey, Andreas Blattmann, Tim Dockhorn, Jonas M{\"u}ller, Joe Penna, and Robin Rombach.
\newblock Sdxl: Improving latent diffusion models for high-resolution image synthesis.
\newblock {\em arXiv preprint arXiv:2307.01952}, 2023.

\bibitem{radford2021learning}
Alec Radford, Jong~Wook Kim, Chris Hallacy, Aditya Ramesh, Gabriel Goh, Sandhini Agarwal, Girish Sastry, Amanda Askell, Pamela Mishkin, Jack Clark, et~al.
\newblock Learning transferable visual models from natural language supervision.
\newblock In {\em International conference on machine learning}, pages 8748--8763. PMLR, 2021.

\bibitem{renduchintala2023tied}
Adithya Renduchintala, Tugrul Konuk, and Oleksii Kuchaiev.
\newblock Tied-lora: Enhacing parameter efficiency of lora with weight tying.
\newblock {\em arXiv preprint arXiv:2311.09578}, 2023.

\bibitem{sap2020commonsense}
Maarten Sap, Vered Shwartz, Antoine Bosselut, Yejin Choi, and Dan Roth.
\newblock Commonsense reasoning for natural language processing.
\newblock In {\em Proceedings of the 58th Annual Meeting of the Association for Computational Linguistics: Tutorial Abstracts}, pages 27--33, 2020.

\bibitem{tian2024hydralora}
Chunlin Tian, Zhan Shi, Zhijiang Guo, Li~Li, and Chengzhong Xu.
\newblock Hydralora: An asymmetric lora architecture for efficient fine-tuning.
\newblock {\em arXiv preprint arXiv:2404.19245}, 2024.

\bibitem{touvron2023llama}
Hugo Touvron, Thibaut Lavril, Gautier Izacard, Xavier Martinet, Marie-Anne Lachaux, Timoth{\'e}e Lacroix, Baptiste Rozi{\`e}re, Naman Goyal, Eric Hambro, Faisal Azhar, et~al.
\newblock Llama: Open and efficient foundation language models.
\newblock {\em arXiv preprint arXiv:2302.13971}, 2023.

\bibitem{valipour2022dylora}
Mojtaba Valipour, Mehdi Rezagholizadeh, Ivan Kobyzev, and Ali Ghodsi.
\newblock Dylora: Parameter efficient tuning of pre-trained models using dynamic search-free low-rank adaptation.
\newblock {\em arXiv preprint arXiv:2210.07558}, 2022.

\bibitem{li2022norm}
Liu Xiaolong, Li~Lujun, Li~Chao, and Anbang Yao.
\newblock Norm: Knowledge distillation via n-to-one representation matching.
\newblock In {\em ICLR}, 2023.

\bibitem{yao2023visual}
Hantao Yao, Rui Zhang, and Changsheng Xu.
\newblock Visual-language prompt tuning with knowledge-guided context optimization.
\newblock In {\em Proceedings of the IEEE/CVF conference on computer vision and pattern recognition}, pages 6757--6767, 2023.

\bibitem{yeh2023navigating}
Shih-Ying Yeh, Yu-Guan Hsieh, Zhidong Gao, Bernard~BW Yang, Giyeong Oh, and Yanmin Gong.
\newblock Navigating text-to-image customization: From lycoris fine-tuning to model evaluation.
\newblock In {\em The Twelfth International Conference on Learning Representations}, 2023.

\bibitem{yu2023task}
Tao Yu, Zhihe Lu, Xin Jin, Zhibo Chen, and Xinchao Wang.
\newblock Task residual for tuning vision-language models.
\newblock In {\em Proceedings of the IEEE/CVF Conference on Computer Vision and Pattern Recognition}, pages 10899--10909, 2023.

\bibitem{zaken2021bitfit}
Elad~Ben Zaken, Shauli Ravfogel, and Yoav Goldberg.
\newblock Bitfit: Simple parameter-efficient fine-tuning for transformer-based masked language-models.
\newblock {\em arXiv preprint arXiv:2106.10199}, 2021.

\bibitem{zanella2024low}
Maxime Zanella and Ismail Ben~Ayed.
\newblock Low-rank few-shot adaptation of vision-language models.
\newblock In {\em Proceedings of the IEEE/CVF Conference on Computer Vision and Pattern Recognition}, pages 1593--1603, 2024.

\bibitem{zhang2023lora}
Longteng Zhang, Lin Zhang, Shaohuai Shi, Xiaowen Chu, and Bo~Li.
\newblock Lora-fa: Memory-efficient low-rank adaptation for large language models fine-tuning.
\newblock {\em arXiv preprint arXiv:2308.03303}, 2023.

\bibitem{adalora}
Qingru Zhang, Minshuo Chen, Alexander Bukharin, Pengcheng He, Yu~Cheng, Weizhu Chen, and Tuo Zhao.
\newblock Adaptive budget allocation for parameter-efficient fine-tuning.
\newblock In {\em The Eleventh International Conference on Learning Representations}, 2023.

\bibitem{zhang2022tip}
Renrui Zhang, Wei Zhang, Rongyao Fang, Peng Gao, Kunchang Li, Jifeng Dai, Yu~Qiao, and Hongsheng Li.
\newblock Tip-adapter: Training-free adaption of clip for few-shot classification.
\newblock In {\em European conference on computer vision}, pages 493--510. Springer, 2022.

\bibitem{zhao2023survey}
Wayne~Xin Zhao, Kun Zhou, Junyi Li, Tianyi Tang, Xiaolei Wang, Yupeng Hou, Yingqian Min, Beichen Zhang, Junjie Zhang, Zican Dong, et~al.
\newblock A survey of large language models.
\newblock {\em arXiv preprint arXiv:2303.18223}, 2023.

\bibitem{zhou2022conditional}
Kaiyang Zhou, Jingkang Yang, Chen~Change Loy, and Ziwei Liu.
\newblock Conditional prompt learning for vision-language models.
\newblock In {\em Proceedings of the IEEE/CVF conference on computer vision and pattern recognition}, pages 16816--16825, 2022.

\bibitem{zhou2022learning}
Kaiyang Zhou, Jingkang Yang, Chen~Change Loy, and Ziwei Liu.
\newblock Learning to prompt for vision-language models.
\newblock {\em International Journal of Computer Vision}, 130(9):2337--2348, 2022.

\bibitem{zhou2020evaluating}
Xuhui Zhou, Yue Zhang, Leyang Cui, and Dandan Huang.
\newblock Evaluating commonsense in pre-trained language models.
\newblock In {\em AAAI}, 2020.

\bibitem{zhu2023prompt}
Beier Zhu, Yulei Niu, Yucheng Han, Yue Wu, and Hanwang Zhang.
\newblock Prompt-aligned gradient for prompt tuning.
\newblock In {\em Proceedings of the IEEE/CVF International Conference on Computer Vision}, pages 15659--15669, 2023.

\bibitem{wei2024auto}
Zimian Zimian~Wei, Lujun~Li Li, Peijie Dong, Zheng Hui, Anggeng Li, Menglong Lu, Hengyue Pan, and Dongsheng Li.
\newblock Auto-prox: Training-free vision transformer architecture search via automatic proxy discovery.
\newblock In {\em AAAI}, 2024.

\end{thebibliography}
\bibliographystyle{plain}

\end{document}